\documentclass{article}

\usepackage[english]{babel}

\usepackage[letterpaper,top=2cm,bottom=2cm,left=3cm,right=3cm,marginparwidth=1.75cm]{geometry}

\usepackage{amsmath}
\usepackage{graphicx}
\usepackage[colorlinks=true, allcolors=blue]{hyperref}

\usepackage{algorithm}
\usepackage{algorithmic}

\usepackage{listings}
\usepackage{xcolor}

\usepackage{amsmath} 

\title{R2VF: A Two-Step Regularization Algorithm to Cluster Categories in GLMs}
\author{Yuval Ben Dror\thanks{Earnix Ltd., Israel; e-mail: yuval.bendror@earnix.com}}

\begin{document}
\maketitle

\begin{abstract}
Over recent decades, extensive research has aimed to overcome the restrictive underlying assumptions required for a Generalized Linear Model to generate accurate and meaningful predictions. These efforts include regularizing coefficients, selecting features, and clustering ordinal categories, among other approaches. Despite these advances, efficiently clustering nominal categories in GLMs without incurring high computational costs remains a challenge. This paper introduces Ranking to Variable Fusion (R2VF), a two-step method designed to efficiently fuse nominal and ordinal categories in GLMs. By first transforming nominal features into an ordinal framework via regularized regression and then applying variable fusion, R2VF strikes a balance between model complexity and interpretability. We demonstrate the effectiveness of R2VF through comparisons with other methods, highlighting its performance in addressing overfitting and identifying an appropriate set of covariates.
\end{abstract}

\section{Introduction}

To illustrate existing methods and emphasize their limitations, we start by considering a linear regression model with categorical features encoded using One-Hot Encoding, and numeric features encoded into binary covariates using a split by percentiles:

\vspace{1em} 
$$
y=\beta_0+\sum_{i=1}^{|N|} \sum_{j=1}^{\left|G N_i\right|} \beta_{n_{i, j}} x_{n_{i, j}}+\sum_{i=1}^{|C|} \sum_{j=1}^{\left|G C_i\right|} \beta_{c_{i, j}} x_{c_{i, j}}+\epsilon 
$$
\vspace{1em} 

With $N=$ set of numeric features, $\mathrm{C}=$ set of categorical features, $G N_i=$ set of dummy-encoded bins for numeric feature number $\mathrm{i}, G C_i=$ set of dummy-encoded bins for categorical feature number $\mathrm{i}, \beta_{n_{i, j}}=$ coefficient for numeric feature i in bin $\mathrm{j}, \beta_{c_{i, j}}=$ coefficient for categorical feature i in bin $\mathrm{j}, \epsilon=$ a random normally distributed error with mean 0 and variance $\sigma^2$.

\noindent A Few Notes Regarding the Initial Model:

\begin{itemize}
    \item The model family is Gaussian, but the discussion could easily be adjusted to Generalized Linear Models \cite{mccullagh2019generalized} by applying the appropriate link function.
    
    \item The numeric features are encoded into binary bins, but the discussion would still apply for standardized linear features or numeric features entered with any other transformation. We assume a split-by-percentiles of a maximum of $n$ bins, with $n$ being the initial threshold. This type of transformation was inspired by \cite{s_fujita_et_al_aglm_2020}. Benefits of the discretization of continuous features are discussed thoroughly in \cite{dougherty1995supervised}.
\end{itemize}

Fitting the linear model after this initial preprocess without regularization could lead to severe overfitting- some features might be irrelevant, and some bins may be too sparse and exhibit high variability. In addition, the number of features could potentially be greater than the number of observations, leading to multi-collinearity and infinitely many solutions. 
    
A well-known and widely used method to address these issues is the Lasso  \cite{tibshirani_regression_1996}. The Lasso penalizes each coefficient using the $L_1$-norm, with a regularization parameter $\lambda$. This approach can shrink certain non-meaningful coefficients to zero.
However, the Lasso still has some drawbacks, especially when applied to one-hot encoded binary covariates. By shrinking coefficients to exactly zero, it essentially merges certain categories with the reference category. This behavior is inadequate because it gives too much weight to the bins selected as the reference levels, and it could potentially unnecessarily weigh down meaningful coefficients. Consider the following example:

Suppose we try to predict $income$ based on $city$. There are 10 different cities represented by the first 10 letters of the alphabet, and for some cities the number of observations is relatively small. City A is the largest, so we use it as the reference level. However, city A is also the city in which the income is highest. Therefore, the Lasso could only merge coefficients of other cities with the rather extreme case of city A. Selecting a different city as the reference, with an income closer to the mean, could improve the validity of the merge, but it would still only allow cities to be merged with the reference city.

\begin{figure}[htbp]
    \centering
    \includegraphics[width=0.8\textwidth]{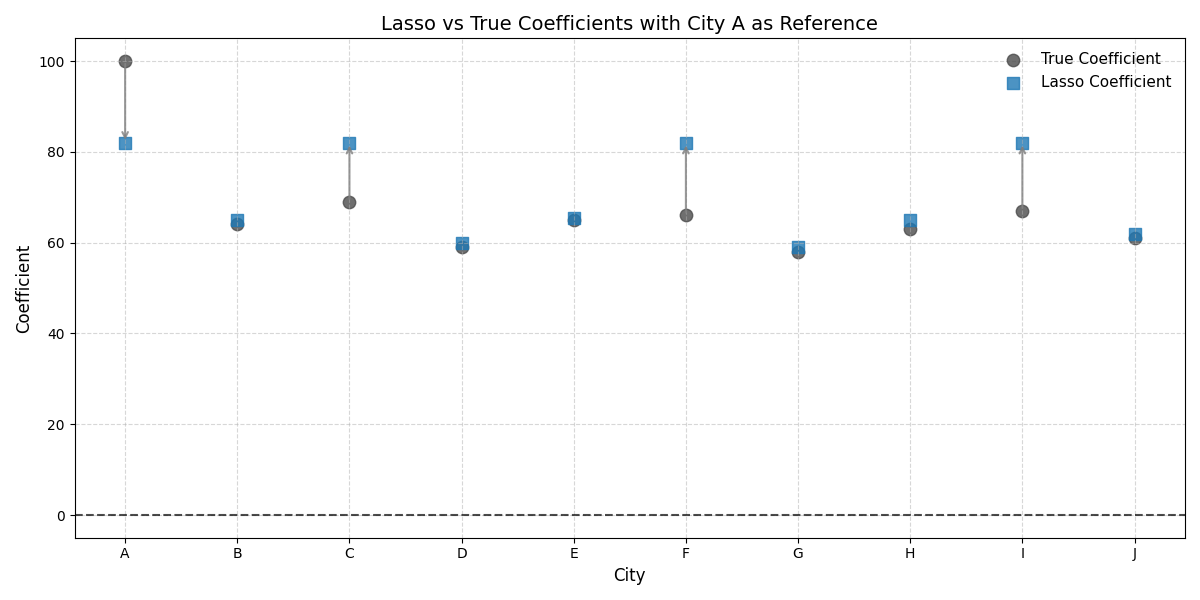} 
    \caption{the x-axis represents the categories of $city$. The y-axis represents the coefficients – the circles are the true coefficients, and the squares are the lasso coefficients. This graph illustrates how some cities are merged with A, thus having their coefficient "dragged upwards".}
    \label{fig:figure_0}
\end{figure}

To address this issue for dummy-encoded numeric features, as well as ordinal categorical features, we could use an alternative called “variable fusion” \cite{land_s_variable_1996}, also commonly known as fused lasso \cite{tibshirani2005sparsity} (which combines standard lasso with variable fusion). This approach uses the following penalty to the coefficients, in case of covariates with an inherent order (such as numeric bins):

\vspace{1em} 
$$
J_\lambda\left(\beta_{\text {ord }}\right)=\lambda \sum_{i=1}^{|O R D|} \sum_{j=1}^{\left|G O_i\right|}\left|\beta_{i j}-\beta_{i j-1}\right|
$$
\vspace{1em} 

With $\lambda=$ a hyper parameter, $\beta_{\text {ord }}=$ the vector of coefficients of ordinal predictors, $O R D=$ the set of ordinal predictors, $G O_i=$ the set of one-hot encoded covariates of ordinal predictor $i$, and $\beta_{i 0}=0$ for all $i$ (which allows us to treat the penalty on the lowest non-reference level as the penalty for its contribution over the reference level).

This penalty effectively merges some of the ordinal bins together by giving them the same coefficient, fusing initial bins to clusters, and leading to a sparser model.

For nominal categorical features, the following penalty was proposed by \cite{bondell_h_d_simultaneous_2009}:

\vspace{1em} 
$$
J_\lambda\left(\beta_{\text {nom }}\right)=\lambda \sum_{i=1}^{|N O M|} \sum_{j<k \in\left|N O_i\right|} w_{i j}\left|\beta_{i j}-\beta_{i k}\right|
$$
\vspace{1em} 

With $\lambda=$ a tuning parameter, $\beta_{\text {nom }}=$ the vector of coefficients of nominal predictors, $N O M=$ the set of nominal predictors, $N O_i=$ the set of one-hot encoded covariates of nominal predictor $i$, and $\beta_{i 0}=0$ for all $i$ (for similar reasons to the ordinal case). Weights $w_{i j}$ can be selected in multiple ways, some of them presented in \cite{gertheiss_sparse_2010}.

However, this approach is computationally expensive, as its implementation involves augmenting the X matrix of $p$ columns to a matrix of $p^2$ columns, before solving an ordinary Lasso problem. When the number of covariates is large, this becomes infeasible in practice, particularly with big datasets.

Therefore, we aim to find a more efficient way to cluster nominal categorical features. In the next section, we present our solution for this challenge. We propose a two-step regularization algorithm, using an initial regularized regression to transform all the nominal categorical features into ordinal features in a regularized framework very similar to our final model, followed by variable fusion. We illustrate the strength of our method in terms of predictive power, sparseness, explainability, and running time. We believe applying this method could be especially useful in environments where explainable and simple models are necessary, and we emphasize the ease-of-use of the solution, limiting the amount of feature-engineering and data-exploration required prior to the fit.

\section{The R2VF Algorithm}

The main idea behind the algorithm is to first establish an empirically-derived order for nominal categories based on their regularized multivariate effects, and then apply variable fusion techniques (typically used for ordinal features) to all features. We first estimate the regularized effects in a model that closely follows the structure of the final variable-fusion model, and then use this knowledge to perform fusion on nominal features. 

\subsection{Outlining the Algorithm}

We call the algorithm “R2VF” – Ranking to Variable Fusion. We will start with a step-by-step description of the algorithm, then explain each step in detail.

\begin{algorithm}
\caption{R2VF Algorithm}
\begin{algorithmic}[1]    
    \STATE Encode categorical variables using One-Hot Encoding, and numeric variables into maximum $n$ bins based on percentiles (with a threshold for minimum observations per bin). \textit{Note: numeric features could also be standardized and entered without binning, or entered with any other transformation.}
    \STATE For numeric features, as well as ordinal categorical features, select the lowest bin as the reference level. For nominal categorical features select the most common category.
    \STATE Use regularized regression with lambda-search on the full list of features – for nominal features, use the standard penalty and for ordinal and numeric features, use the variable fusion penalty (\cite{land_s_variable_1996}). \textit{Note: in this stage, we could use Lasso or Ridge regularization – we discuss the benefits of each method in the next subsection.}
    \STATE 	For each nominal feature, use the regularized coefficients found in Step 3 to encode categories into numbers (with the coefficient for the reference levels being 0). Perform a split by percentiles on the transformed column with maximum $m$ bins.
    \STATE For all features, select the lowest bin as the reference level.
    \STATE Use Lasso variable fusion with lambda-search, applying the penalty on adjacent bins for numeric and ordinal features, and adjacent grouped categories (sorted in Step 4) for categorical features. In the resulting model, treat the categories and bins with the same coefficient as “clusters” with an identical effect over the target.
    \STATE \textit{(Optional)} re-fit the model with the clusters found in step 6, without regularization. 

\end{algorithmic}
\end{algorithm}

\begin{figure}[htbp]
    \centering
    \includegraphics[width=0.9\textwidth]{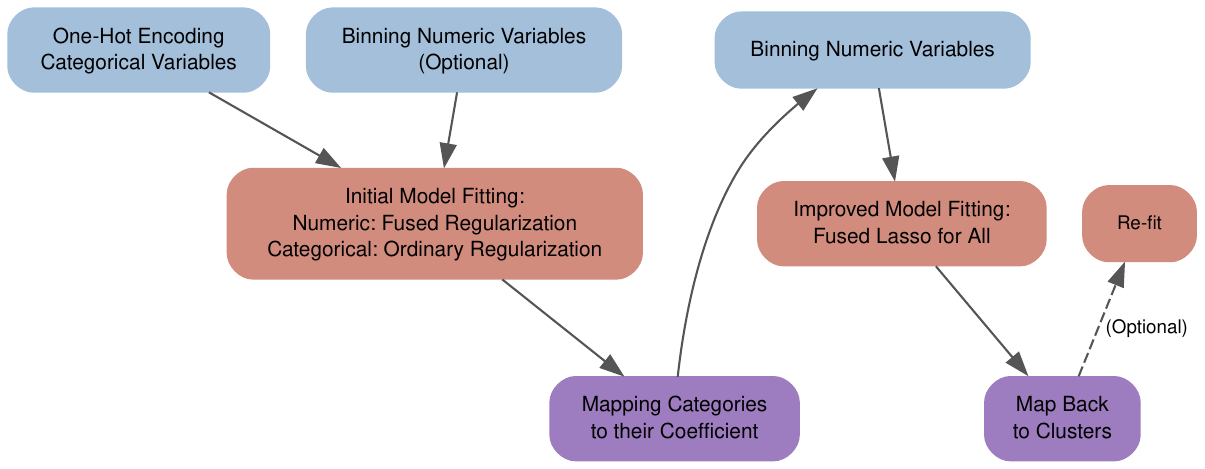}
    \caption{Overview of the R2VF Algorithm Flow}
    \label{fig:r2vf_schema}
\end{figure}

In Step 1, we use a standard one-hot encoding for categorical features, and a split into $n$ bins for numeric features. Different methods could be applied to decide the value of $n$, but they should all involve making sure there is a sufficient number of observations per bin - see further discussion at the next subsection. The main idea here is to capture non-linear relationships between numeric features and the target, exploiting the capabilities of variable fusion in merging adjacent bins. However, this transformation of numeric features is optional; the rest of the algorithm would still apply without it.

In Step 2, for numeric and ordinal features, we use the lowest bin as the reference to make sure the penalty for the second-lowest bin is simply the penalty for its contribution over the reference bin. For categorical features, we select the most common category as our reference. Alternatively, we could also use a category where the mean of the target is close to the overall mean, as discussed in the introduction. 

In Step 3, for numeric and ordinal features we use variable fusion, which helps regularize for the effect of the difference between adjacent bins. This helps us get a close approximation of the final model. For nominal features, we use standard regularization. At this stage, we could use either the Ridge \cite{hoerl1970ridge} or the Lasso penalty– we discuss it at the end of the section. In any case, we get an approximation of the multivariate regularized effect each category has on the overall predictions of a similarly structured model. 

\vspace{1em} 
$$
J_\lambda(\beta) = \lambda \left(\sum_{i=1}^{|ORD|} \sum_{j=1}^{|G O_i|} \left|\beta_{o_{i, j}} - \beta_{o_{i, j-1}}\right|^\alpha 
+ \sum_{i=1}^{|NOM|} \sum_{j=1}^{|N O_i|} \left|\beta_{c_{i, j}}\right|^\alpha \right)
$$

\vspace{1em} 

In this penalty formula, $ORD$ represents the set of ordinal features (numeric and categorical), and $NOM$ represents the set of nominal categorical features. $\alpha$ could be either equal to 1 or 2, as discussed above.

In Step 4, we use the coefficients from Step 3 to transform the categorical variables to numeric variables in a way that considers the multivariate effect of the category in a similar model. Note that if the number of categories is larger than $m$, some categories will already be merged in this stage through the percentile-based split on the coefficients. See discussion regarding the value for $m$ at the end of the section. 

In Step 5, like in Step 2, we ensure that the penalty of the first coefficient in the variable fusion is for its contribution over the reference bin.

In Step 6, we use variable fusion across all features. This is allowed because now all the bins can be ordered meaningfully. That means each coefficient in the model is penalized with regards to the “previous coefficient”– for numeric and ordinal features the “previous” means the previous in the natural numeric (or logical) order, and for nominal features that means the category with the closest multivariate effect. This ensures a similar regularization environment across all bins.

\vspace{1em} 
$$
J_\lambda(\beta)=\lambda\left(\sum_{i=1}^{|O R D|} \sum_{j=1}^{\left|G O_i\right|}\left|\beta_{O_{i, j}}-\beta_{O_{i, j-1}}\right|\right)
$$
\vspace{1em} 

Here, all features belong to the set $ORD$.

In Step 7, we use the clusters found in Step 6 to perform a non-regularized re-fit. This is an optional step, inspired by the tests conducted by \cite{gertheiss_sparse_2010} which show that it could help in avoiding over-shrinkage of coefficients of covariates that actually go into the model.

\subsection{A Note On Hyper Parameters Selection}

Two of the hyper parameters of the model created by the algorithm remain, for now, heuristic and configurable by the user:

\begin{itemize}
    \item The selection of $n$ (number of initial bins for numeric and ordinal feature) and $m$ (number of initial bins for the transformed categorical features)- $n$ is not particularly relevant for this paper's discussion, because as mentioned before, the split of the numeric features is mostly done to illustrate the effectiveness of variable fusion. The selection of $n$ is discussed at \cite{s_fujita_et_al_aglm_2020}. For $m$, things are a bit more tricky. The most obvious option is to simply give an initial bin for each category. However, from our experiments, in cases of high cardinality features this can lead to massive over-fitting even for relatively high lambdas, causing the optimal lambda found by the grid-search to be ultimately too high. This can be partially solved by using lasso, not ridge, in step 3, but a more beneficial approach would be to select an $m$ closer to the value of $n$, with the reasonable assumption that the average number of levels of effect of categorical features shouldn't be much higher than the average number of levels of effect of numeric features. In the insurance industry, in which we operated to create the algorithm, a good value for $m$ was usually somewhere between 50 and 100, and a good value for $n$ was usually around 30.
    
    \item   The selection of Lasso vs Ridge for the ranking step (Step 3) - the advantage of using Lasso is that we gain a model that resembles the final model more closely. The disadvantage is that it performs a ranking where some categories are already merged with the reference level. Ridge gives a full ranking, but it’s not as similar to the final model as the Lasso. From our experiments, there was no consistent winner between the two options.
    
\end{itemize}

\section{Computational Approach}

We will describe the computational implications of our algorithm by showing that it doesn't involve customizing the known regularizations, nor inflating $X$ or $Y$. If we treat $X$ and $Y$ as constants and mark the running time as $O\left(L_\lambda\right)$ for ordinary Lasso regression \cite{friedman2010regularization} with a lambda-search \cite{hastie2009elements}, and $O\left(R_\lambda\right)$ for Ridge regression, then the run-time of our algorithm is $O\left(L_\lambda\right)+O\left(R_\lambda\right)$ in case we use Ridge in Step 3, or 2$O\left(L_\lambda\right)$ in case we use Lasso in Step 3. None of the steps require optimizing more than one factor, nor implementing complex penalties, such as the one presented by \cite{bondell_h_d_simultaneous_2009}.

For variable fusion we use split-coding \cite{walter_coding_1987}, then apply an ordinary Lasso penalty. The idea behind split-coding is that the bin-covariates: $b_i$ ( $i=$ left bin edges) for a value C in the original ordinal feature are defined by:

\vspace{1em} 
$$
\begin{aligned}
b_i & =1 \text { for } C \geq i \\
b_i & =0 \text { otherwise }
\end{aligned}
$$

Therefore, the model coefficients are parameterized by:

\vspace{1em} 
$$
\delta_i=\beta_i-\beta_{i-1}
$$

So, transitions between bin $b_i$ and $b_{i-1}$ are expressed by coefficient $\delta_i$, and to retrieve $\beta_i$ we only need to use the back-transformation $\beta_i=\sum_{s=1}^i \delta_s$. This approach basically performs regularization on the contribution of each bin over the previous bin- identical to the variable fusion Lasso. There is no inflation of parameters with this scheme.

We note that some adjustments could be made to the computation to decrease computation time– using one validation set instead of cross-validation, especially for the ranking step, could be beneficial, considering the fact that we’re only interested in the order of the categories and not the coefficients themselves. Alternatively, we could use only a sample of the data table for the ranking step. In terms of the lambda grid, certain optimizations could be used in order to search on a smaller grid in Step 6, especially if we use Lasso in Step 3– but these are beyond the scope of this paper.

\section{Comparisons of Model Performance and Sparseness}

To execute the R2VF algorithm in Python, we perform data manipulations as a preprocess and a post-process as outlined in Sections 2 and 3, and use the H2O Python package to create the GLMs with lambda searches. After finding the optimal covariates we run a non-regularized GLM (re-fit). To do the lambda search we use a validation set (the same set for Step 3 and Step 6). In these examples, we performed a Lasso regularization in Step 3.

We start with a simulated dataset in which certain categories are generated with the same coefficients. First we will illustrate how the algorithm works, and then we’ll perform some comparisons of model performance. After that, we will continue with an example on a real dataset.

\subsection{Simulated Data}

We start by creating 3 variables which are somewhat correlated with each other, as well as a target column, as follows:

\begin{description}
    \item[\textbf{City:}] 26 cities labeled A to Z, randomly generated such that the number of observations per city roughly forms a linear scale (meaning, the frequency of each city varies),
    \item[\textbf{Age:}] An average age is randomly selected per city (varies from 34 to 46), and generated with a variability of 13.
    \item[\textbf{Profession:}] Marked \( P_i \) (where \( i \) is a number from 0 to 99), and distributed such that it has a minor correlation with both city and age. The distribution makes some professions relatively prevalent, some very rare, and others completely absent.
\end{description}

\begin{description}
\item[\textbf{Target:}] Generated using the following formula (introducing some variability as well)-
\end{description}

\begin{lstlisting}[caption={Code: Synthetic Dataset}, label={lst:synthetic_dataset}]
target = 0 \
    + 15 * (row['city'] in ['b', 'c', 'd']) \
    + 17 * (row['city'] in ['e', 'f', 'g', 'h', 'i', 'j', 'k', 'l']) \
    - 12 * (row['city'] in ['m', 'n', 'o', 'p']) \
    - 14 * (row['city'] in ['q', 'r', 's', 't', 'u', 'v']) \
    + 10 * (row['city'] in ['w', 'x']) - 10 * (row['city'] in ['y', 'z']) \
    - 2 * np.sqrt((row['age'] - 45))**2 - 19 * (row['profession'][-1] == '1') \
    - 17 * (row['profession'][-1] == '2') - 9 * (row['profession'][-1] == '3') \
    - 8 * (row['profession'][-1] == '4') + 1 * (row['profession'][-1] == '5') \
    + 2 * (row['profession'][-1] == '6') + 8 * (row['profession'][-1] == '7') \
    + 9 * (row['profession'][-1] == '8') + 19 * (row['profession'][-1] == '9')
\end{lstlisting}

Here, some of the cities have an identical coefficient based on alphabetical order, and some of the professions have an identical coefficient based on their last digit. The age is modeled as a V-shape centered at 45 (to illustrate the effectiveness of numeric binnings). 

We start with a graphical illustration of the R2VF algorithm. For simplicity, we will focus on the $city$ variable, but it’s important to emphasize the fact that it’s part of a multivariate model.

\begin{figure}[h!]
    \centering
    \includegraphics[width=0.8\textwidth]{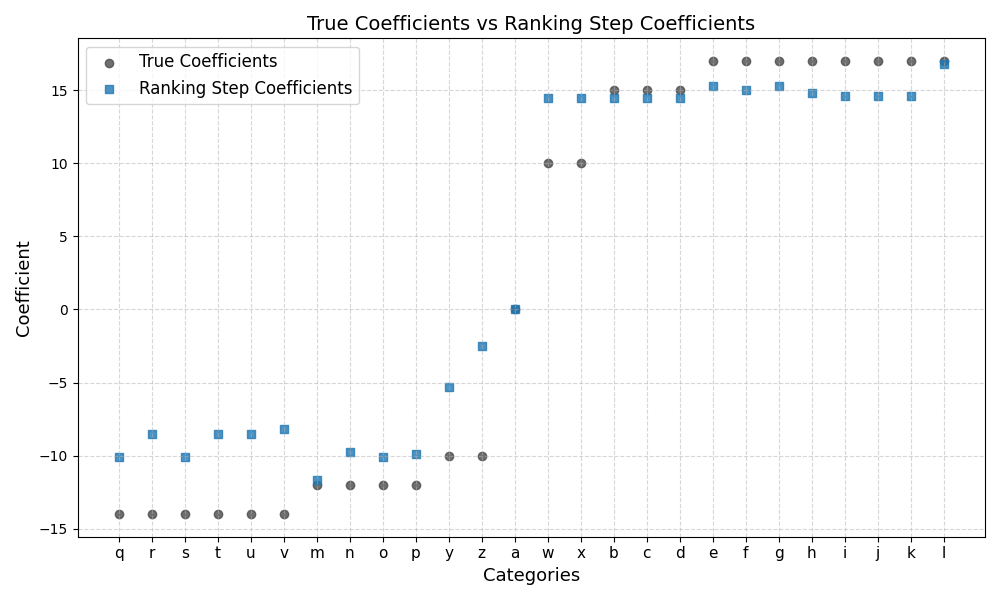} 
    \caption{the x-axis represents the categories of $city$ ordered by their true coefficient. The y-axis represents the coefficients – the circles are the true coefficients, and the squares are the coefficients given by the Step-3 model (the ranking step), after fixing the reference level to be “a” by adding a constant value.}
    \label{fig:figure_1}
\end{figure}

We run the algorithm on the simulated dataset. In Step 3, we get a ranking of the categories on the regularized environment. We note that category “b” was selected as the initial intercept. In this step, since we used Lasso as the ranking algorithm, some categories were already merged with “b”. The Figure~\ref{fig:figure_1} graph illustrates how an ordinary Lasso regularizes towards one value (in this case, the value for “b”), which causes over-shrinkage of the lower values upwards, and over-shrinkage of the higher values downwards. However, the ranking itself is still close to the real desired ranking in the multivariate environment.

\begin{figure}[h!]
    \centering
    \includegraphics[width=0.8\textwidth]{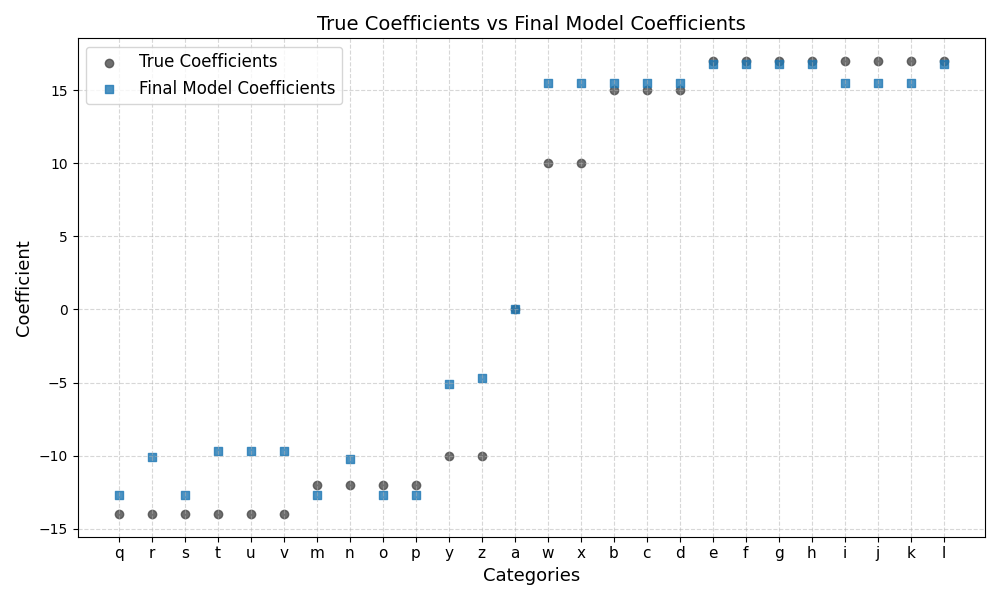} 
    \caption{the x-axis represents the categories of $city$ ordered by their true coefficient. The y-axis represents the coefficients – the circles are the true coefficients, and the squares are the coefficients given by the final model. Note that some categories got the same coefficient, meaning they were merged to the same cluster.}
    \label{fig:figure_2}
\end{figure}

In the next step, we encode the categories to the coefficient given to them as shown in Figure~\ref{fig:figure_1}. Next, we use variable fusion to cluster certain categories together. The final coefficients are given in Figure~\ref{fig:figure_2}. We note that the 26 cities were merged to 9 clusters (compared to 7 “real” clusters). In this case, it is apparent that the regularization (and subsequently, the clusters themselves) shrank coefficients towards each other rather than to a specific value– in some cases, the real value was slightly lower than the true coefficient, and in some cases it was slightly higher. This illustrates the effectiveness of the regularization obtained by clustering categories with each other.

\subsection{Real Data}

This dataset is taken from the FARS (Fatality Analysis Reporting System) annual file of 2022– we use the vehicle dataset, containing vehicle data for car accidents in 2022 in US. (\cite{nhtsa_fars_2022})

We will create a binary model predicting the probability of an accident resulting in a death.

3 of the 5 predictors are categorical – $STATENAME$ refers to the state in which the accident took place, $MAK\_MOD$ refers to the model of the vehicle, and $BODY\_TYP$ refers to the body type of the vehicle. $MONTH$ and $HOUR$ refer to the time of the accident. 
Number of unique categories: $STATENAME$ – 51, $MAK\_MOD$  – 886, $BODY\_TYP$ - 59. 

We perform a cross-testing to get a distribution of metrics on test data. We split the data 5 times to get 5 separate sets of training-test. We perform a comparison of the following models:

\begin{description}
\item[\textbf{R2VF:}] implemented as explained in the beginning of the section.
\item[\textbf{OLVF:}] Ordinary Lasso for categorical features + variable fusion Lasso for Numeric Features. 
\item[\textbf{No Regularization:}] using the initial bins as the final covariates without regularization.
\item[\textbf{Catboost of Main Effects:}] a Catboost \cite{prokhorenkova2018catboost} model with default settings and an overfit detector, set to maximum depth 1 (to avoid interactions, which are irrelevant for this discussion). 

\end{description}

\begin{figure}[h!]
    \centering
    \includegraphics[width=0.8\textwidth]{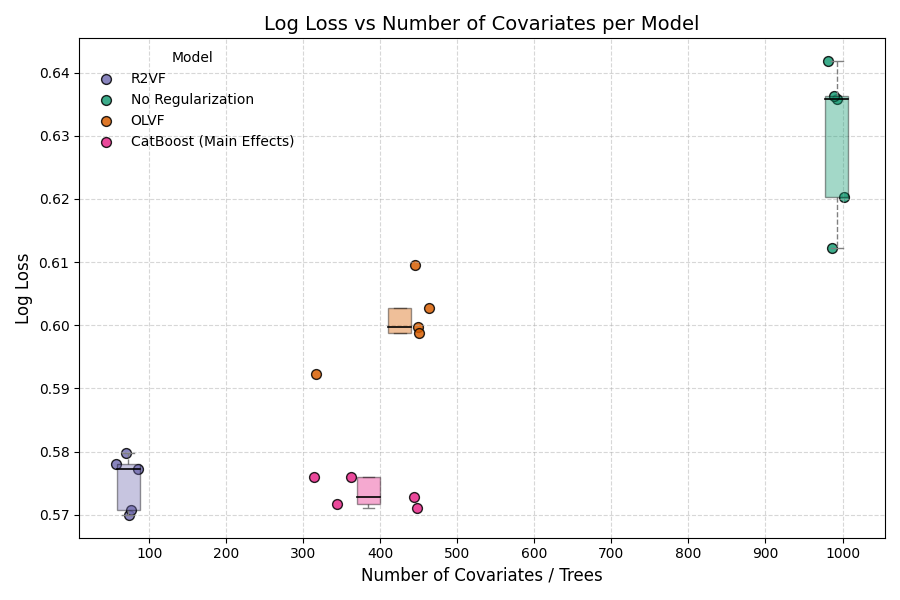} 
    \caption{Log-Loss on test data for 5 separate sets of train-test splits. The x-axis represents the number of covariates used in the final model – for Catboost, it means the number of trees created (since its depth is 1). The boxplots are located on the mean value of covariates used.}
    \label{fig:figure_4}
\end{figure}

According to the results, presented in Figure~\ref{fig:figure_4}, the performance of R2VF in terms of log-loss across datasets is on-par with that of the Main Effects Catboost, and far surpasses the performance of the ordinary Lasso with variable fusion.  The complexity of the R2VF model is much smaller than the other models – the number of covariates is over 3 times smaller than the ordinary Lasso, as well as the number of Catboost trees. As for Catboost - we consider it to be a a state-of-the-art benchmark for categorical predictions in a noisy environment, as it combines different methods of target encoding \cite{micci2001preprocessing}, regularization and cross-folding. We note that while Catboost effectively handles categorical features (e.g., through ordered target statistics), it does not explicitly cluster categories in an interpretable way that maps groups of original categories to a single coefficient.

\section{Discussion}

The R2VF algorithm provides much sparser models than current solutions, with an improvement on model performance as well. Despite these results, some areas are still ripe for inquiry. A pivotal aspect concerns the impact of reference level selection during the initial regularized regression step. Although our methodology primarily leverages this step for category ranking, minimizing its overall impact, the selection could skew the regularization strength across categories. Specifically, categories with coefficients closer to the intercept might undergo less regularization compared to those further away. Investigating methods to mitigate this bias could yield a more robust category ranking process. 

Another challenge we face is determining the regularization type in Step 3 – Ridge regression has the advantage of resulting in a full ranking which doesn’t merge any categories with the intercept, but Lasso regression is much closer to the final model and thus probably provides a more accurate ranking for this specific model. Initial results suggest that while Ridge in Step 3 provides a full ranking, for complex models (high-cardinality or many features), this dense ranking (when binned in Step 4) might create an overly granular set of ordinal levels. The subsequent Lasso variable fusion in Step 6 might then select a very high lambda to compensate, potentially leading to over-regularization of the final clusters.

In addition, computation time could be further optimized, particularly during the initial regularized regression stage. Given our focus on category ranking, rather than being interested in the actual coefficients, various strategies could streamline the process—utilizing a subset of data, refining the lambda search grid, or simplifying the validation approach. It could be beneficial to explore these and other heuristics to shorten computation time without significantly affecting the category ranking accuracy.

Finally, we note that integrating the elastic net, as introduced by \cite{zou_regularization_2005}, into the regularization steps could further refine the model, offering a more nuanced balance between both the Ridge and Lasso benefits. This modification could potentially enhance the model's flexibility and applicability across a broader range of datasets and scenarios.

\section*{Acknowledgments}

The author would like to express gratitude to Earnix Ltd. for supporting this research and providing the resources necessary for its completion. This work was conducted as part of the author’s role at Earnix, focusing on the development of Auto-GLM. A special thanks goes to the author's supervisors during the development period, Luba Orlovsky, Boris Pritsker, Reuven Shnaps and Erez Barak, as well as to the teammates during this period who provided valuable insights - Shilo Horev, Tzviel Frostig, Yitzhak Yahalom and Eyal Bar Natan.

\bibliographystyle{plain} 
\bibliography{references} 

\begin{thebibliography}{10}

\bibitem{bondell_h_d_simultaneous_2009}
{BONDELL, H. D.} and {REICH, B. J.}
\newblock Simultaneous factor selection and collapsing levels in anova.
\newblock In {\em Biometrics 65}, pages 169--177. 2009.

\bibitem{dougherty1995supervised}
James Dougherty, Ron Kohavi, and Mehran Sahami.
\newblock Supervised and unsupervised discretization of continuous features.
\newblock In {\em Machine learning proceedings 1995}, pages 194--202. Elsevier, 1995.

\bibitem{friedman2010regularization}
Jerome~H Friedman, Trevor Hastie, and Rob Tibshirani.
\newblock Regularization paths for generalized linear models via coordinate descent.
\newblock {\em Journal of statistical software}, 33:1--22, 2010.

\bibitem{gertheiss_sparse_2010}
Jan Gertheiss and Gerhard Tutz.
\newblock Sparse modeling of categorial explanatory variables.
\newblock {\em The Annals of Applied Statistics}, 4(4):2150--2180, December 2010.
\newblock Publisher: Institute of Mathematical Statistics.

\bibitem{hastie2009elements}
Trevor Hastie, Robert Tibshirani, Jerome~H Friedman, and Jerome~H Friedman.
\newblock {\em The elements of statistical learning: data mining, inference, and prediction}, volume~2.
\newblock Springer, 2009.

\bibitem{hoerl1970ridge}
Arthur~E Hoerl and Robert~W Kennard.
\newblock Ridge regression: Biased estimation for nonorthogonal problems.
\newblock {\em Technometrics}, 12(1):55--67, 1970.

\bibitem{land_s_variable_1996}
{Land, S.} and {Friedman, J.}
\newblock Variable fusion: a new method of adaptive signal regression.
\newblock 1996.

\bibitem{mccullagh2019generalized}
Peter McCullagh.
\newblock {\em Generalized linear models}.
\newblock Routledge, 2019.

\bibitem{micci2001preprocessing}
Daniele Micci-Barreca.
\newblock A preprocessing scheme for high-cardinality categorical attributes in classification and prediction problems.
\newblock {\em ACM SIGKDD explorations newsletter}, 3(1):27--32, 2001.

\bibitem{nhtsa_fars_2022}
{NHTSA}.
\newblock {FARS} 2022, 2022.

\bibitem{prokhorenkova2018catboost}
Liudmila Prokhorenkova, Gleb Gusev, Aleksandr Vorobev, Anna~Veronika Dorogush, and Andrey Gulin.
\newblock Catboost: unbiased boosting with categorical features.
\newblock {\em Advances in neural information processing systems}, 31, 2018.

\bibitem{s_fujita_et_al_aglm_2020}
{S Fujita et al.}
\newblock {AGLM}: {A} {Hybrid} {Modeling} {Method} of {GLM} and {Data} {Science} {Techniques}.
\newblock 2020.

\bibitem{tibshirani_regression_1996}
Robert Tibshirani.
\newblock Regression {Shrinkage} and {Selection} {Via} the {Lasso}.
\newblock {\em Journal of the Royal Statistical Society: Series B (Methodological)}, 58(1):267--288, January 1996.

\bibitem{tibshirani2005sparsity}
Robert Tibshirani, Michael Saunders, Saharon Rosset, Ji~Zhu, and Keith Knight.
\newblock Sparsity and smoothness via the fused lasso.
\newblock {\em Journal of the Royal Statistical Society Series B: Statistical Methodology}, 67(1):91--108, 2005.

\bibitem{walter_coding_1987}
S.~D. WALTER, A.~R. FEINSTEIN, and C.~K. WELLS.
\newblock {CODING} {ORDINAL} {INDEPENDENT} {VARIABLES} {IN} {MULTIPLE} {REGRESSION} {ANALYSES}.
\newblock {\em American Journal of Epidemiology}, 125(2):319--323, February 1987.

\bibitem{zou_regularization_2005}
Hui Zou and Trevor Hastie.
\newblock Regularization and {Variable} {Selection} {Via} the {Elastic} {Net}.
\newblock {\em Journal of the Royal Statistical Society Series B: Statistical Methodology}, 67(2):301--320, April 2005.

\end{thebibliography}

\end{document}